\pdfoutput=1

\documentclass[11pt]{article}

\usepackage[table]{xcolor} 
\usepackage[final]{acl}
\usepackage{tabularx}%
\usepackage{multirow}
\usepackage{times}
\usepackage{latexsym}
\usepackage{booktabs}
\usepackage{arydshln}

\usepackage{hyperref}
\usepackage[T1]{fontenc}

\usepackage[utf8]{inputenc}

\usepackage{microtype}

\usepackage{inconsolata}

\usepackage{graphicx}
\usepackage{multirow}
\usepackage{subfig}
\usepackage{color}
\usepackage{colortbl}
\usepackage{multirow}

\makeatletter
{\small 
\xdef\f@size@small{\f@size}
\xdef\f@baselineskip@small{\f@baselineskip}
\normalsize 
\xdef\f@size@normalsize{\f@size}
\xdef\f@baselineskip@normalsize{\f@baselineskip}
}
\newcommand{\smalltonormalsize}{%
  \fontsize
    {\fpeval{(\f@size@small+\f@size@normalsize)/2}}
    {\fpeval{(\f@baselineskip@small+\f@baselineskip@normalsize)/2}}%
  \selectfont
}
\makeatother

\title{On the Benchmarking of LLMs for Open-Domain Dialogue Evaluation}

\author{John Mendonça\textsuperscript{1,2},  Alon Lavie\textsuperscript{3,4} \and Isabel Trancoso\textsuperscript{1,2} \\
  \textsuperscript{1} INESC-ID, Lisbon \\
  \textsuperscript{2} Instituto Superior Técnico, University of Lisbon \\
  \textsuperscript{3} Carnegie Mellon University, Pittsburgh \\
  \textsuperscript{4} Phrase, Pittsburgh \\
  \texttt{\{john.mendonca, isabel.trancoso\}@inesc-id.pt}, \texttt{alavie@cs.cmu.edu} \\}

\begin{document}
\maketitle
\begin{abstract}
Large Language Models (LLMs) have showcased remarkable capabilities in various Natural Language Processing tasks. For automatic open-domain dialogue evaluation in particular, LLMs have been seamlessly integrated into evaluation frameworks, and together with human evaluation, compose the backbone of most evaluations. However, existing evaluation benchmarks often rely on outdated datasets and evaluate aspects like \textit{Fluency} and \textit{Relevance}, which fail to adequately capture the capabilities and limitations of state-of-the-art chatbot models. 

This paper critically examines current evaluation benchmarks, highlighting that the use of older response generators and quality aspects fail to accurately reflect modern chatbot capabilities. A small annotation experiment on a recent LLM-generated dataset (SODA) reveals that LLM evaluators such as GPT-4 struggle to detect actual deficiencies in dialogues generated by current LLM chatbots.


\end{abstract}

\section{Introduction}

The emergence of Large Language Models has brought about a significant transformation in the field of Natural Language Processing (NLP). Through advancements in model and data scale and clever prompting strategies that leverage In Context Learning and Chain-of-Thought reasoning \citep{wei2022chain}, these models have exhibited remarkable capabilities, marking a notable shift in the landscape of NLP research and development.

For the task of dialogue evaluation in particular, several works have proposed the use of LLMs as a standalone or as part of a dialogue evaluation framework for direct assessment of a response or full dialogue \citep{liu-etal-2023-g,mendonca-etal-2023-simple,lin-chen-2023-llm}. By simply providing inside the prompt the quality aspects under evaluation, together with a few examples, these models achieve state of the art correlations with human judgements on several benchmarks.

\begin{table}[t]
\smalltonormalsize
\centering
\begin{tabularx}{0.47\textwidth}{ X }
\toprule
\textbf{Context:}\\
\textbf{A:} I am chatting with clients on the internet.\\
\textbf{B:} What kind of service are you providing them?\\\midrule
\textbf{Mitsuku Response:} Is there only one? \\
\textbf{Fluency:} 5 \textbf{Relevance:} 2 \textbf{Overall Score:} 2 \\ \midrule
\textbf{Claude-3 Response:} Ah, well you see, I'm what you might call a... digital concierge, of sorts. My role is to engage with clients online, understand their needs and preferences and provide them with personalised assistance and recommendations.\\
\textbf{Fluency:} 5 \textbf{Relevance:} 5 \textbf{Overall Score:} 5 \\ \midrule
\textbf{LLama3 Response:} I'm offering technical support and answering questions about software products.\\
\textbf{Fluency:} 5 \textbf{Relevance:} 5 \textbf{Overall Score:} 4 \\ \bottomrule
\end{tabularx}
\caption{Responses given example context obtained from the FED benchmark. Responses are generated by benchmark's chatbot (Mitsuku), a contemporary closed-source (Claude-3-haiku) and open-source (Llama-3-70B-instruct) chatbot. Scores provided by GPT-4.}
\label{tab:sampleshowcase}
\end{table}

\begin{table*}[t]
\scriptsize
\begin{tabular}{cccccc}
\toprule
Annotation           & Dataset                        & Type      & Lang     & Quality Aspects                                                                                                                                                                 & Generation Models                                                                                \\ \midrule
\multirow{2}{*}{FED} & \begin{tabular}[c]{@{}c@{}}Meena, Mitsuku,\\ Human-Machine\end{tabular} & Turn      & EN       & \begin{tabular}[c]{@{}c@{}}Interesting, Engaging, Specific, Relevant,\\ Correct, Semantically Appropriate, Understandable,\\ Fluent, Overall\end{tabular}                       & \multirow{2}{*}{Human, Meena, Mitsuku}                                                           \\
                     &                                & Dial      & EN       & \begin{tabular}[c]{@{}c@{}}Coherent, Recover, Consistent, Diverse, Depth,\\ Likeable, Understanding, Flexible, \\ Informative, Inquisitive, Overall\end{tabular}                &                                                                                                  \\ \midrule
\multirow{2}{*}{USR} & PersonaChat                    & Turn      & EN       & \multirow{2}{*}{\begin{tabular}[c]{@{}c@{}}Understandable, Natural, Maintains Context,\\ Interesting, Uses Knowledge, Overall\end{tabular}}                                     & \multirow{2}{*}{\begin{tabular}[c]{@{}c@{}}Transformer, Seq2Seq,\\ LSTM,KV-MemNN\end{tabular}}   \\
                     & TopicalChat                    & Turn      & EN       &                                                                                                                                                                                 &                                                                                                  \\ \midrule
DSTC10               & Mixture                        & Turn & EN       & \begin{tabular}[c]{@{}c@{}}Appropriateness, Content,\\ Grammatical, Relevance\end{tabular}                                                                                                                                                                       & \begin{tabular}[c]{@{}c@{}}LSTM, HRED, BlenderBot,\\ DialoGPT, T5, GPT-3\end{tabular}                    \\ \midrule
DSTC11               & Mixture                        & Turn+Dial & EN,ES,ZH & \begin{tabular}[c]{@{}c@{}}Appropriateness, Content Richness,\\ Grammatical Correctness, Relevance, Coherence, \\ Engageness/Likeability, Informativeness, Overall\end{tabular} & \begin{tabular}[c]{@{}c@{}}DSTC10, GPT-3.5, ChatGPT,\\ BlenderBot3, Xiaoice, PlatoXL\end{tabular} \\ \bottomrule
\end{tabular}
\caption{Human annotation benchmarks used to evaluate LLM-based open-domain dialogue evaluators.}
\label{tab:benchmarks}
\end{table*}

Despite the promising results heralded by this recent approach, we argue that the methods used to benchmark dialogue evaluation are not adequate to accurately assess the evaluation capabilities of current open-domain dialogue evaluation metrics. 

In this paper, we investigate existing commonly used human-annotated datasets and identify their shortcomings when used as benchmarks for assessing LLM-based evaluators. In particular, these datasets often rely on the use of weak chatbots to evaluate the proposed framework/metric (as illustrated in Table \ref{tab:sampleshowcase}). Consequently, the commonly probed quality aspects have as a primary focus issues such as \textbf{Fluency} (\textit{Is the response written correctly?}) and \textbf{Relevance} (\textit{Is the response relevant given the context?}). With the introduction of LLMs, the evaluation of these aspects is rendered mostly useless. Yet, existing benchmarks continue to prioritise these outdated criteria, leading to a disconnect between evaluation practices and the capabilities of modern chatbots. 

In support of our argument, we present a small qualitative analysis of evaluations provided by these models on dialogues that better approximate current chatbot performance. On the one hand, our analysis shows that dialogues that lack \textit{Fluency} are both easy to detect, and hard to find. On the other hand, LLMs struggle to correctly identify \textit{Coherence} and \textit{Commonsense} issues, which are aspects where the current generation of chatbots still under-perform and where better detection and evaluation would be desirable.

With these contributions, we seek to highlight the following:

\textbf{1. There is an urgent need for new and more meaningful benchmarks}. In particular, the release of more human annotations of responses and dialogues generated by contemporary LLMs is necessary to provide a better benchmarking framework for new evaluation methodologies.

\textbf{2. Evaluation methodologies must be informed by current chatbot capabilities.} Open-domain evaluation should focus on identifying novel frontiers in dialogue generation. We argue that aspects such as \textit{Coherence} and \textit{Commonsense} should take the forefront in evaluation instead of \textit{Fluency} or \textit{Relevance}.

\section{Benchmark datasets}
\label{sec:datasets}

This section presents a brief survey of datasets that have been used as a benchmark for LLM-based open-domain dialogue evaluation metrics. These datasets are summarised in Table \ref{tab:benchmarks} for ease of reference.

The \textbf{FED dataset \citep{mehri-eskenazi-2020-unsupervised}} consists of turn level and dialogue level annotations of conversations conducted between a Human (40 dialogues) and two chatbot engines (\textbf{Meena} with 40 dialogues \citep{DBLP:journals/corr/abs-2001-09977} and 44 from \textbf{Mitsuku} \footnote{\href{https://medium.com/pandorabots-blog/mitsuku-wins-loebner-prize-2018-3e8d98c5f2a7}{Mitsuku blogpost}}) targeting eighteen quality aspects. Each conversation received one annotation at the dialog level and three annotations at the turn level, randomly selected from the conversation. In total, the FED dataset comprises 3,348 turn-level and 1,364 dialog-level data points, amounting to 4,712 annotations.

For \textbf{USR \citep{mehri-eskenazi-2020-usr}}, annotations were collected for models trained on the TopicalChat \citep{Gopalakrishnan2019} and PersonaChat \citep{zhang-etal-2018-personalizing} dialogue datasets. Generated responses were obtained from models including \textbf{Transformer} \citep{46201}, \textbf{RNN Seq2Seq} \citep{shang-etal-2015-neural}, \textbf{LSTM} \citep{10.1162/neco.1997.9.8.1735}, and \textbf{KV-MemNN} \citep{miller-etal-2016-key}. For each dialog context, an additional human response was also collected. Human annotation was then carried out on sixty dialog contexts, with six responses per context for Topical-Chat (four transformer outputs with different decoding strategies, one newly-annotated human output, and the original ground-truth response) and five for PersonaChat (Seq2Seq, LSTM, KV-MemNN, one newly-annotated human output, and the original ground-truth response).

\textbf{DSTC10 \citep{zhang2021automatic}.} The principal goal of the "Automatic Evaluation and Moderation of Open-domain Dialogue Systems" track was to offer a competitive venue for participants in this challenge to design robust automatic dialogue evaluation metrics that correlate well with human judgements across multiple dialogue domains as well as across different quality aspects. For the development set, 14 publicly available datasets were collected: (1-3) GRADE Datasets \citep{huang-etal-2020-grade}, (4-5) DailyDialog/Persona-Zhao \citep{zhao-etal-2020-designing}, (6) DailyDialog-Gupta \citep{gupta-etal-2019-investigating}, (7-8) USR, (9) HUMOD \citep{app10030762}, (10) Twitter-DSTC6 \citep{hori2018endtoend}, (11) Reddit-DSTC7 \citep{Galley2019GroundedRG}, (12) Persona-See \citep{see-etal-2019-makes} and (13-14) FED. In total, over 35k turn-level human annotations were compiled. For testing, 3 sources of data were used: (1) CHANEL-JSALT2020, (2) ChatEval
\citep{sedoc-etal-2019-chateval} and (3) an additional annotation conducted on TopicalChat \citep{Gopalakrishnan2019} and PersonaChat \citep{zhang-etal-2018-personalizing}. Eight systems, a human baseline, and a random utterance were used as response generators. Specifically, the eight systems are \textbf{LSTM Seq2Seq}, \textbf{Attention-based LSTM Seq2Seq} \citep{10.5555/2969033.2969173}, \textbf{HRED} \citep{10.5555/3016387.3016435}, \textbf{VHRED}, \textbf{BlenderBot (400M-Distill)} \citep{roller-etal-2021-recipes}, \textbf{DialoGPT-medium} \citep{zhang-etal-2020-dialogpt}, \textbf{T5-base} \citep{10.5555/3455716.3455856}, and \textbf{GPT-3} \citep{brown2020language}.

\textbf{DSTC11 \citep{rodriguez-cantelar-etal-2023-overview}.} Similar to DSTC10, the "Robust and Multilingual Automatic Evaluation Metrics for Open-Domain Dialogue Systems" track is split into development and test sets. For the development set, the organisers provide data from two clusters of datasets from DSTC10 and 4,470 dialogues (approximately 130k turns) open-domain human-human dialogues which are originally in Chinese. Since the goal of the shared task was to evaluate mulitlinguality and robustness of metrics, development data is translated into English, Chinese, Spanish, and back-translated. For testing, the organisers combine a portion of the DSTC10 test set, and include new Human-Chatbot dialogues generated by SotA chatbots. These are: \textbf{ChatGPT} (a platform powered by GPT-3.5-Turbo), \textbf{GPT-3.5} \citep{ouyang2022training} and \textbf{BlenderBot3} \citep{shuster2022blenderbot3A}. Similar to the development set, the test set was also translated. In total, 4,839 turn level and 277 dialogue level annotations were conducted.

\section{LLMs as evaluators}
\label{sec:llms}

Most automatic evaluation in the literature up until recently was conducted with word-overlap metrics or encoder-based metrics trained using self-supervised training objectives \cite{yeh-etal-2021-comprehensive}. \citet{mehri-eskenazi-2020-unsupervised} proposed an alternative approach called \textbf{FED} (fine-grained evaluation of dialog), which measures dialogue quality by computing the likelihood that DialoGPT \citep{zhang-etal-2020-dialogpt} will respond to it with a particular set of follow-up utterances that are constructed.

Despite the unsupervised nature, it was only with the introduction of LLMs that these approaches fully replaced encoder-based metrics. 

The first documented systematic evaluation of LLMs was conducted by \citet{huynh2023understanding}, where they evaluate training and few-shot strategies for this task. The authors evaluate several LLMs including BLOOM \citep{workshop2023bloom}, OPT \citep{zhang2022opt}, GPT-3, Flan-T5 \citep{chung2022scaling}, InstructDial \citep{gupta-etal-2022-instructdial} and TNLGv2 \citep{smith2022using} on the \textbf{DSTC10} and \textbf{FED} benchmarks. The authors report good correlation results with human judgements and confirm the appropriateness of few-shot learning for dialogue evaluation.

\textbf{GPTScore \citep{fu2023gptscore}} is based on the assumption  that a generative pre-training model will assign a higher probability to high-quality generated text than low quality one following a given instruction and context. Several LLMs are tested, including GPT-3 and Flan-T5 on the \textbf{FED-turn} dataset. 

\textbf{G-Eval \citep{liu-etal-2023-g}} studies GPT-3.5-Turbo and GPT-4 for the evaluation of generation models. In detail, the framework comprises (1) a prompt defining the evaluation task and criteria, (2) a Chain-Of-Thoughts step containing intermediate instructions generated by the LLM outlining evaluation steps, and (3) a scoring function based on return token probabilities estimated by generating multiple times. For the task of dialogue evaluation, G-Eval is benchmarked on the \textbf{USR-TopicalChat} dataset covering naturalness, coherence, engagingness and groundedness.

\textbf{DialEvalML \citep{mendonca-etal-2023-simple}} is a hybrid framework combining encoder-based models (in this case XLM-RoBERTa-large \citep{conneau-etal-2020-unsupervised}) trained with self-supervised objectives and direct prompting and score extraction from GPT-3.5-Turbo. The authors combine the predictions using a correlation rescaling method obtained from the development set, 
achieving first place in all tracks of \textbf{DSTC11} \citep{rodriguez-cantelar-etal-2023-overview}.

\textbf{LLM-Eval \citep{lin-chen-2023-llm}} is a single-prompt-based evaluation method that leverages a unified evaluation schema to cover multiple dimensions of conversation quality in a forward pass. The authors evaluate Claude-v1.3 \cite{claude}, ChatGPT and GPT-3.5 on the \textbf{DSTC10} hidden set.

\textbf{XDial-Eval \citep{zhang-etal-2023-xdial}} focuses on probing the evaluation capabilities of several open access LLMs against GPT-3.5-Turbo. The authors focus on context relevance and coherence by combining a selection of subsets from \textbf{DSTC11} development set. They additionally translate the original English data 
to 9 additional languages. Unlike other approaches, the LLMs were evaluated in (1) zero and few shot learning scenario; (2) instruction finetuning; and (3) ensemble with a strong encoder-based framework.

\textbf{\citet{zhang2024comprehensive}} conduct a comprehensive study of 30 recently emerged LLMs for automatic dialogue evaluation using a smaller subset than the one from XDial-Eval. In particular, the authors assess \textit{Relevance}, \textit{Understandability}, \textit{Specificity}, \textit{Interestingness}, and \textit{Overall quality} at the turn level, while at the dialogue level, they evaluate \textit{Coherence}, \textit{Engagingness}, \textit{Informativeness}, \textit{Diversity}, and \textit{Overall quality}.

\section{Limitations in Current Benchmarking}
\label{sec:limitations}

Given the datasets identified in Section \ref{sec:datasets} used to assess LLM-based evaluators (Section \ref{sec:llms}), we identify several limitations in the benchmarking of automatic open-domain dialogue evaluation, which we enumerate below.

\paragraph{Use of Outdated Generative Models}

With the exception of DSTC11-test (which was only used as a benchmark by \texttt{DialEvalML}), most benchmarks contain responses from older generative models such as LSTMs or HRED. As a result, a substantial amount of low quality responses are easily identifiable (lacking basic quality aspects such as fluency, contextual relevance or specificity). Concurrently, responses that are relevant but contain contradictions, coherence issues or are factually incorrect are overvalued by evaluators due to biased guidelines. This tendency to rate flawed responses can skew the perception of evaluator performance, leading to misleading conclusions about their effectiveness in practice. 

\paragraph{Irrelevance of Quality Aspects in Current Chatbots}

Dialogue evaluation guidelines are focused on detecting issues that were prevalent in older generation models. For instance, all benchmarks have a quality aspect that targets \textit{Fluency} and \textit{Relevance}. Given current LLM-based chatbots, these quality aspects are no longer informative to differentiate output quality between different contemporary dialogue systems: most if not all models now output fluent and relevant responses (e.g., Table \ref{tab:sampleshowcase}).

\paragraph{Focus on English}

An overarching trend on the benchmarks being used is that they exclusively focus on the English language. Although DSTC11 does provide annotations in Chinese and Spanish, they are only partially available for the test set. Moreover, in the development set, only translated versions of the original English dialogues are included, thereby introducing English bias into the assessment of quality. This bias further extends to the test set, where, even if evaluated by native annotators, the aspects being measured fall short of capturing the linguistic and cultural nuances present in dialogues. These nuances can include the use of formal versus informal language, expressions of politeness, cultural references, and idiomatic expressions\footnote{Visit \href{culturalatlas.sbs.com.au/about}{Cultural Atlas} for a centralised repository of various cultures and corresponding communication practices.} that may not directly translate into English.

\section{Qualitative Analysis}
\label{sec:experiment}

Informed by the issues highlighted in Section \ref{sec:limitations}, we conduct a small scale annotation experiment. The goal of this annotation is twofold. Firstly, we aim to understand whether annotations such as \textit{Fluency} are still relevant. Secondly, the annotation of more complex aspects such as \textit{Coherence} or \textit{Commonsense} in this dataset allows us to understand the performance of LLMs when evaluating responses generated by SoTA chatbots on quality aspects that require a stronger understanding of conversational dynamics. 

\begin{figure*}
\centering
\includegraphics[width=0.49\textwidth]{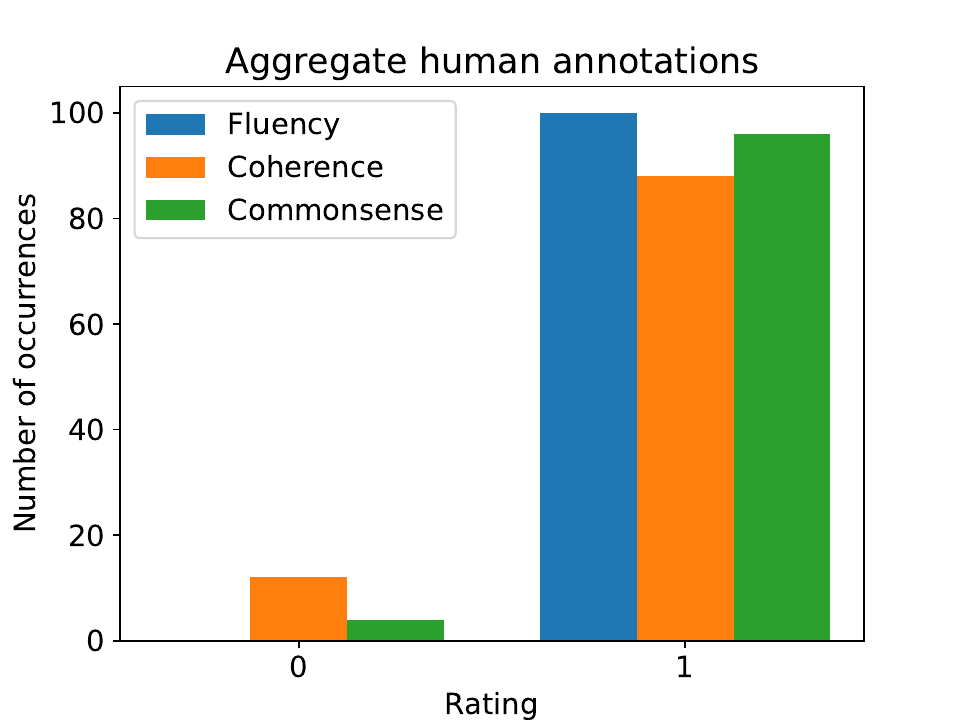}
\includegraphics[width=0.49\textwidth]{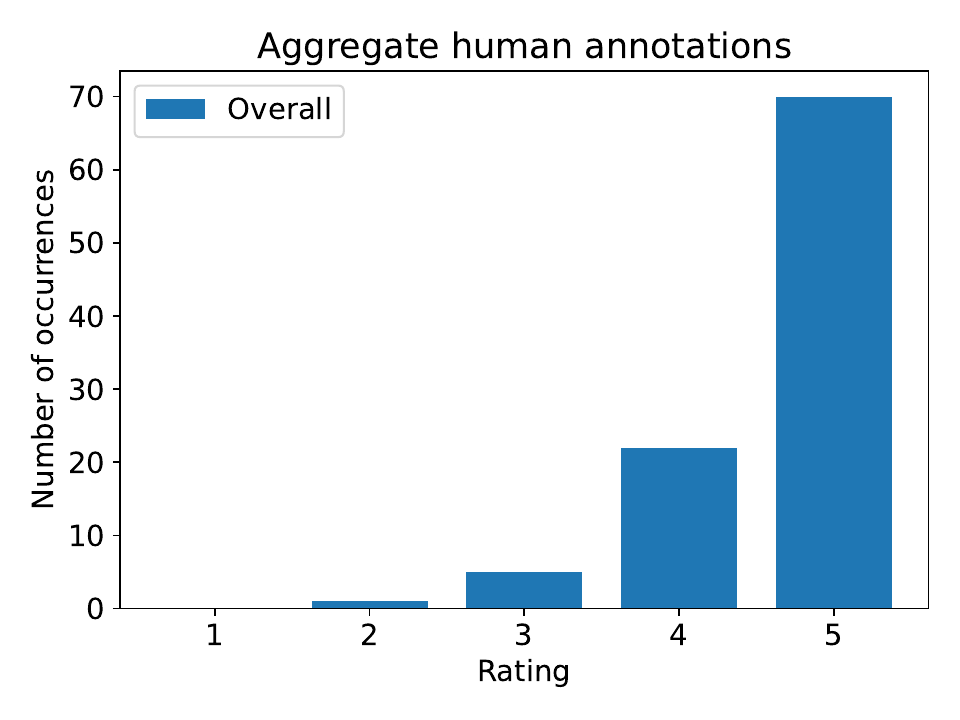}
\caption{Aggregate human annotations on SODA. Annotations for \textit{Overall} rounded down to the nearest integer.}
\label{fig:annot}
\end{figure*}

We use SODA \citep{kim-etal-2023-soda} as our dialogue dataset since it leverages a LLM (in this case GPT-3.5) for the generation of dialogues. As such, SODA will exhibit most of the typical issues associated with LLMs, thereby making its use as a contemporary benchmark more relevant than benchmarks relying on weaker response generators (as identified in Section \ref{sec:limitations}).
Human evaluation conducted on SODA shows that its dialogues are more consistent, specific, and natural than DailyDialog \citep{li-etal-2017-dailydialog}, a frequently used dialog dataset used for the development of evaluation metrics \citep{yeh-etal-2021-comprehensive}. Table \ref{tab:sodaexample} presents an example of the SODA dataset, where a \textit{Coherence} issue is highlighted.

\subsection{Annotation}

We recruited 3 expert annotators \footnote{All annotators are members of our research lab.} to rate the first 100 dialogues\footnote{The evaluated dialogues have a turn distribution similar to the one of the full SODA dataset (average of 4 turns per dialogue, minimum 2 and maximum 8).} of the test set in terms of:

\begin{itemize}
    \item \textit{Fluency (0,1):} The dialogue is written correctly and has no grammatical errors.
    \item \textit{Coherence (0,1):} The dialogue is coherent and does not contain contradictions within itself.
    \item \textit{Commonsense (0,1):} The dialogue does not contain common sense issues. It is logical, makes sense and is aware of basic facts and effects.
    \item \textit{Overall quality [1,5]:} Overall impression of the dialogue.
\end{itemize}

\begin{table}[h]
\centering
\begin{tabular}{lc}
\toprule
Aspect      & Spearman \\ \midrule
Fluency     & -        \\
Coherence   & 0.7025   \\
Commonsense & 0.6534   \\
Overall     & 0.7425  \\\bottomrule
\end{tabular}
\caption{Inter annotator agreement for each aspect studied. All correlations p\textless0.05.}
\label{tab:iaa}
\end{table}

\begin{table}[ht]
\small
\centering
\begin{tabularx}{0.48\textwidth}{ X }
\vspace{0.2cm}
\cellcolor[HTML]{eeeeee}Your task is to evaluate dialogues in terms of Fluency, Coherence, Commonsense and Overall Quality.\\\vspace{0.2cm}
\cellcolor[HTML]{eeeeee}Fluency (0-bad,1-good): The dialogue is written correctly and has no grammatical errors.\\\vspace{0.001cm}
\cellcolor[HTML]{eeeeee}Coherence (0-bad,1-good): The dialogue is coherent and does not contain contradictions within itself. E.g.: Someone saying they are flying to London for the first time and then saying they went there before in a subsequent turn.\\\vspace{0.001cm}
\cellcolor[HTML]{eeeeee}Commonsense (0-bad,1-good): The dialogue does not contain common sense issues. It is logical, makes sense and is aware of basic facts and effects. E.g. Drinking a coffee as a refreshment for the summer lacks commonsense.\\\vspace{0.001cm}
\cellcolor[HTML]{eeeeee}Overall (1 (poor) up to 5 (excellent)): Overall impression of the dialogue.\\\vspace{0.2cm}
\cellcolor[HTML]{eeeeee}Please present your evaluation into the following json format:\\
\cellcolor[HTML]{eeeeee}\{"Fluency": \_, "Coherence": \_, "Commonsense": \_, "Overall": \_\}\\\vspace{0.2cm}
\cellcolor[HTML]{eeeeee}Dialogue:\\
\cellcolor[HTML]{eeeeee}\hspace{0.18\textwidth}\textbf{[Dialogue]}
\vspace{0.2cm}
\end{tabularx}
\caption{Dialogue evaluation instruction template (denoted as \textit{Ours} in the experiments).}
\label{tab:prompt}
\end{table}

\begin{table*}[t]
\centering
\begin{tabular}{lcccc}
\toprule
Evaluator     & Fluency (Acc.) & Coherence ($r_{pb}$) & Commonsense ($r_{pb}$) & Overall ($\rho$) \\ \midrule
G-EVAL 3.5 \citeyearpar{liu-etal-2023-g}   & 0.99                        & 0.2283                  & \textit{0.0425}                   & 0.2716                \\
G-EVAL 4     & 0.97                        & 0.1749                   & \textbf{0.4348}                    & 0.3789                \\\midrule
LLM-EVAL 3.5 \citeyearpar{lin-chen-2023-llm}& \textbf{1.00}                           & 0.1834                  & 0.1993                    & 0.2435                \\
LLM-EVAL 4   &\textbf{1.00}                           & 0.2489                  & 0.4054                    & \textbf{0.3811}                \\\midrule\midrule
Ours GPT-3.5     & 0.99                        & 0.2721                   & 0.3353                      & 0.1857                \\
Ours GPT-4       & 0.99                        & 0.1659                   & 0.3440                    & 0.3782  \\\midrule
Ours Llama-3-8B     & 0.99                & 0.1155                   & \textit{0.0205}                      & 0.1953                \\
Ours Llama-3-70B     & 0.99                & \textbf{0.2722}                   & \textit{0.0205}                      & 0.2115                \\\bottomrule
\end{tabular}
\caption{Evaluation results with human judgements on SODA. Performance for \textit{Fluency} is reported using Accuracy, \textit{Coherence} and \textit{Commonsense} using Point-biserial correlation and \textit{Overall} with Spearman correlation. \textbf{Bold} denotes best performance. All correlations p\textless0.05 unless \textit{italicised}.}
\label{tab:main_Results}
\end{table*}

Following \citet{mehri-eskenazi-2020-unsupervised}, we report inter annotator agreement results in Table \ref{tab:iaa}, corresponding to the correlation between each annotation and the mean of the annotations for the same quality aspect. For \textit{Fluency}, all annotators reported 0 dialogues with issues. As such, the correlation (and most other agreement metrics) is undetermined. For the other annotations, agreement is high, and in line with other works (\citet{mehri-eskenazi-2020-unsupervised} reports correlations as low as 0.562 for \textit{Consistency}.). Figure \ref{fig:annot} presents the aggregate annotations for the SODA dataset. These aggregate ratings are computed using majority voting for the binary aspects and simple average (rounded down) for \textit{Overall}.

With respect to the annotations that target specific aspects of quality, the majority of dialogues were annotated as \textit{fluent}, \textit{coherent} and with \textit{commonsense}. In particular, the annotations did not identify \textbf{any} \textit{Fluency} issues in all dialogues. This supports our argument that annotating \textit{Fluency} has limited value given current chatbot capabilities. 

\subsection{Baseline Evaluators}

As a baseline for the analysis, we evaluate two typically used closed-source LLMs: GPT-3.5-Turbo and GPT-4 \footnote{\texttt{gpt-3.5-turbo-0125} and \texttt{gpt-4-1106-preview} accessed via OpenAI's API in early April.}, using the prompting strategies of G-Eval \citep{liu-etal-2023-g}, LLM-EVAL \citep{lin-chen-2023-llm}, and our own contribution. Additionally, we probe the performance of Llama-3 \citep{llama3modelcard}, an open access model with benchmark performances \footnote{\href{github.com/meta-llama/llama3/blob/main/eval\_details.md}{\texttt{Llama-3} reported evaluation}} similar to the closed source ones:

\begin{itemize}
    \item \textbf{G-Eval} calculates an average score sampled from 20 generations with high temperature. We obtain a binary decision for \textit{Fluency} when s \textgreater \space0.5. 
    \item \textbf{LLM-EVAL} outputs a score from 1-100. Similar to G-Eval, we consider a dialogue to be fluent when s \textgreater \space50.
    \item \textbf{Our contribution} directly probes the LLM using the same guidelines provided to the annotators, therefore the scores are extracted directly. The template used is presented in Table \ref{tab:prompt}. 
\end{itemize}

We provide in the prompt the full dialogue and ask the LLM to rate the dialogue according to the probed aspects. We follow the hyperparameters of the original work whenever available. For our method, we employ a \textit{temperature} of 0.3 for GPT models and 0.6 for Llama, and generate a single output.

For evaluation, we employ metrics adapted to the aggregate labels. For \textit{Fluency}, since all dialogues are rated as being fluent, we use simple accuracy; for \textit{Coherence} and \textit{Commonsense}, we report results using point-biserial correlation ($r_{pb}$) since the labels provided are binary (0,1); finally, \textit{Overall} results are presented using Spearman ($\rho$) correlation (1-5 Likert score).

\subsection{Results}

We present the evaluation results for our annotated subset in Table \ref{tab:main_Results}.

\paragraph{Fluency} With the exception of LLM-EVAL, all evaluators failed to correctly identify all dialogues as being fluent. One dialogue in particular contains a hallucination that affects the understanding of the dialogue, but is still strictly fluent. As such, the performance of LLM-EVAL can be attributed to the 0-100 scoring scale, which allows for a more fine grained evaluation of the dialogue. In fact, LLM-EVAL outputs a much lower score (still above the decision threshold of 50) to this dialogue when compared to other ones. In any case, we consider this to be an edge case of a failed evaluation that could be resolved by providing a more comprehensive prompt and/or including examples. 

\begin{figure*}
\centering
\includegraphics[width=\textwidth]{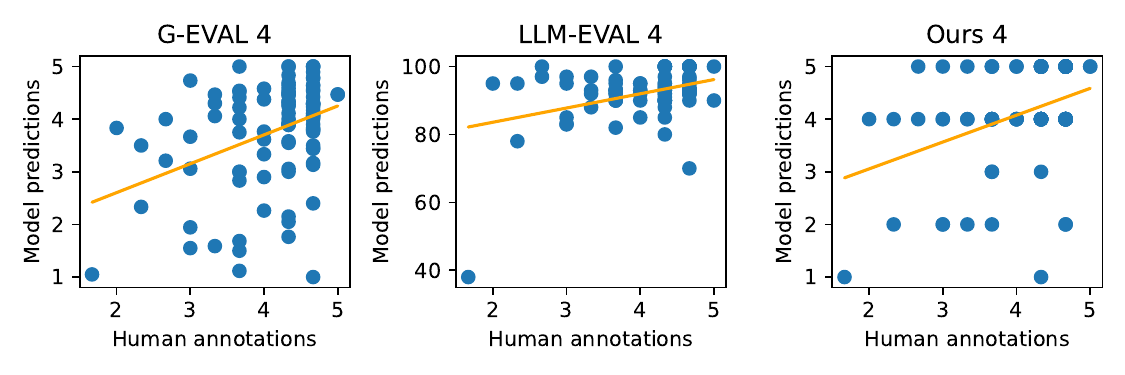}
\caption{Scatter plots and corresponding correlations for each prompting strategy using GPT-4.}
\label{fig:scatters}
\end{figure*}

\paragraph{Coherence} Generally speaking, LLM evaluators struggle with correctly identifying responses that lack \textit{Coherence}, with the best approaches only achieving \textbf{.2722} correlation (LLama-3-70B). 
Using our prompting strategy, we note that these approaches were only able to correctly classify 1 (GPT-3.5-Turbo) and 2 (GPT-4) out of 12 incoherent dialogues, underlining the difficulty these models have in identifying coherence issues. In fact, GPT-3.5-Turbo only rated a single dialogue as lacking coherence (against the 6 dialogues rated by GPT-4), which explains why it has larger correlations than GPT-4 (lower false positives). Table \ref{tab:sodaexample} presents an example of such failed prediction.

\paragraph{Commonsense} When compared to \textit{Coherence}, LLMs have much larger variability in performance for \textit{Commonsense}. For instance, GPT-4 achieves over .4 correlation using G-Eval and LLM-EVAL prompting strategies, whereas the LLama-3 model evaluations and G-EVAL 3.5 are mostly uncorrelated. The low score for LLama-3 could be attributed to a difficulty in understanding the evaluation task. Given GPT-3.5 worked reasonably well for the other prompting strategies, the performance is explained by the disagreement between individual sampled scores. In any case, the predictions are generally better for \textit{Commonsense} than with \textit{Coherence}, which could be explained by the fact that illogical actions can be mostly identified directly, and without taking into account prior contextual details found in the dialogue. This contrasts with \textit{Coherence}, which requires a deeper contextual analysis to detect inconsistencies, which is a known limitation of LLMs \citep{han2023information}.

\begin{table}[ht]
\small
\centering
\begin{tabularx}{0.48\textwidth}{ X }
\toprule
\cellcolor[HTML]{eeeeee}\textbf{A:} I've been thinking a lot lately about \underline{moving back to}
\underline{my home country.} \\\addlinespace
\cellcolor[HTML]{e8f4f8}\textbf{B:} Really? Why?\\\addlinespace
\cellcolor[HTML]{eeeeee}\textbf{A:} I miss my family a lot. And I want to be closer to
them as they get older. I can provide support and
assistance to them more easily if I am living nearby.\\\addlinespace
\cellcolor[HTML]{e8f4f8}\textbf{B:} I can understand that. But what about us?\\\addlinespace
\cellcolor[HTML]{eeeeee}\textbf{A:} We can still visit each other. \underline{And it's not like we're}
\underline{moving to different countries.} We'll still be in the
same region.\\\addlinespace
\cellcolor[HTML]{e8f4f8}\textbf{B:} True, But it's going to be a big adjustment for both of us.\\\addlinespace
\cellcolor[HTML]{eeeeee}\textbf{A:} I know it will be tough at first, but I think it will be
worth it in the long run. Plus, you could always come
\underline{visit me in my home country!}\\\addlinespace
\cellcolor[HTML]{e8f4f8}\textbf{B:} Hmmm...I don't know if I'm ready for that kind of
commitment just yet. But I'm willing to try it if you are.\\ \addlinespace\midrule\midrule\addlinespace
Human annotation:\\
\textbf{Fluency:} 1 \textbf{Coherence:} 0 \textbf{Commonsense:} 0 \textbf{Overall:} 2\\ \midrule
Ours GPT-4:\\
\textbf{Fluency:} 1 \textbf{Coherence:} 1 \textbf{Commonsense:} 1 \textbf{Overall:} 5\\
\bottomrule
\end{tabularx}
\caption{Example dialogue extracted from SODA, together with Human and GPT-4 rating. The underline text identifies a coherence issue.}
\label{tab:sodaexample}
\end{table}

\paragraph{Overall} Similar to \textit{Commonsense}, \textit{Overall} predictive performance is stronger when using GPT-4 as the base LLM evaluator, with the best correlations being achieved using LLM-EVAL 4 at \textbf{.3811}. Nevertheless, this correlation rate is still subpar when compared against reported dialogue-level correlations on other benchmark datasets -- for instance, LLM-EVAL reports a 0.71 correlation on \textbf{FED-dialogue} (\textit{Overall Quality}). Figure \ref{fig:scatters} presents scatter plots for GPT-4 predictions across the probed prompting strategies.

\subsection{Discussions}

\paragraph{Model size}

Overall, we note that the larger models (GPT-4 vs GPT-3.5, LLama-3-70B vs LLama-3-8B) consistently outperform their corresponding smaller models for both Coherence and Commonsense. This may be attributed to breakthrough performance thanks to model scaling, which has also been reported as "emergent abilities" in complex reasoning tasks \citep{52065}. This observation contrasts with Fluency, where no difference has been noted between model size.

\paragraph{External Expert Knowledge} Surprisingly, we find instances where the model considers a high quality dialogue to be low quality. Upon further inspection, these ratings appear to have been influenced by external expert knowledge, something the annotators did not take into account. For instance, in one of the dialogues, one of the participants is asking for advice to patent a catalytic converter they invented. This is picked up by the evaluator when asked for an explanation: \textit{"there is a significant commonsense issue: the catalytic converter is not a new invention."}. This is an incorrect evaluation within the framework of our study since it is not commonsense knowledge. Nevertheless, this topic is of significant interest for evaluation and is not explicitly studied in many benchmarks. In fact, it might be one type of evaluation LLMs can excel at, especially when individual annotator knowledge is limited.

\paragraph{Dialogue length}

\begin{figure}[t]
\centering
\includegraphics[width=0.5\textwidth]{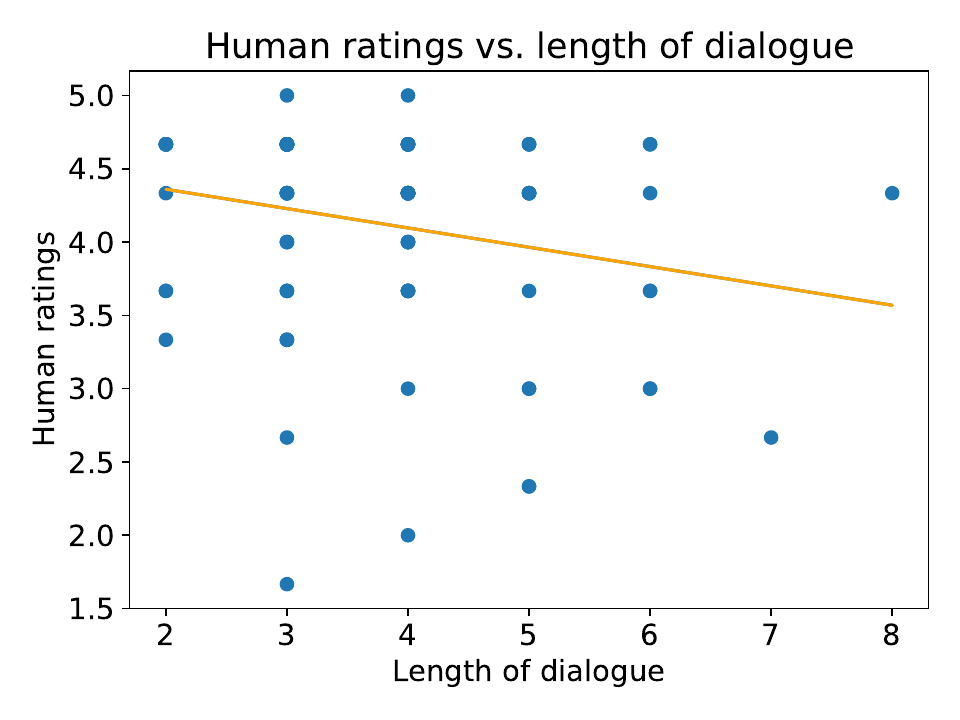}
\caption{Scatter plots of human ratings against dialogue length.}
\label{fig:length}
\end{figure}

The limitations of LLM reasoning and understanding over long contexts is well documented in the literature \citep{bai2024longbench,kuratov2024babilong}. As such, one possible reason for issues in the dialogue could be attributed to dialogue length. With this in mind, we calculate the Point-biserial correlation ($r_{pb}$) between \textit{Coherence}/\textit{Commonsense} and the length of the dialogue. For \textit{Coherence}, we report a correlation of -0.228, which denotes a small to medium correlation; for \textit{Commonsense}, correlation is non-significant (0.006). We additionally present the scatter plot for \textit{Overall} in Figure \ref{fig:length}. Similarly to \textit{Coherence}, we report a Spearman correlation of -0.251. Firstly, as expected, commonsense issues are mostly independent to dialogue length, which makes sense since commonsense knowledge is drawn from model training and not from context. For coherence, its correlation with dialogue length is small. However, we acknowledge that the small sample size of larger dialogues does not allow for more definitive conclusions.

\section{Conclusion}

This paper conducts an inventory of open domain dialogue evaluation benchmarks being currently used by LLM evaluation frameworks. We show that these benchmarks have several limitations that hinder the progress in the field. In particular, we argues they lack (1) responses generated by strong LLM chatbots; (2) aspects that identify their weaknesses; (3) representation of other languages and cultures. In order to illustrate these limitations, we also conducted a small scale experiment on SODA and show that even GPT-4 shows limitations in the detection of low quality responses. 

However, these findings underscore one critical limitation in how direct assessment benchmarks are currently being developed: they are mostly concerned with evaluating contemporary chatbot capabilities. As it stands, the current evaluation research environment is one where the driver of progress is the advancement in generation, and not the converse. Ultimately, evaluation benchmarks should possess the flexibility to remain relevant as newer chatbots emerge, thereby pushing the envelope of dialogue generation. Embracing this goal would not only foster greater comparability and reproducibility in research, but also facilitate continuous improvement in the field, leading to the development of better chatbots. 


\section{Limitations}

\paragraph{Pairwise Comparisons} Our study is focused on metrics that predict human judgements on singular responses or dialogues. We acknowledge other methodologies such as pairwise comparisons exist, and that they mostly circumvent the limitations we highlight. Nevertheless, given the documented interest in the literature of metrics that are optimised to predict direct assessments provided by humans, we argue our study is still valuable. Furthermore, direct assessments provide a more granular assessment of response quality that pairwise comparisons lack, especially when comparing models that differ only slightly in quality but are otherwise similar \citep{smith-etal-2022-human}.

\paragraph{SODA} Unlike the majority of benchmarks studied, where chatbots generate a response given seed human-human interaction or conducts a full conversation with a human, SODA dialogues are entirely synthetic. As such, one might argue this approach may hide possible limitations of chatbots since they are in control of the whole conversation, thereby excluding human feedback within the conversation which can be used to aid evaluation \citep{petrak-etal-2023-learning}. However, there are very few open source open-domain dialogue datasets that contain LLMs as one of the participants\footnote{In fact, most recent user-LLM chatbot interaction datasets are conversational QA \citep{zheng2024lmsyschat1m, zhao2024wildchat}.}.

\paragraph{Self-evaluation biases}

One consideration in the current LLM-based evaluation paradigm is that self-evaluation biases may arise. This bias is more evident in subjective assessments such as "Overall Quality", which is more pronounced in pairwise comparisons \citep{panickssery2024llm}. While this bias can be reduced by employing more objective quality aspects such as the ones we propose in this work, it is still possible that models will erroneously overlook their own errors. As such, it is important to complement automated direct assessment with human judgements.

\paragraph{Monolingual} We identified English-centric evaluation as one the issues in current benchmarking. However, our experiment is conducted on SODA, which is exclusively in English. The aim of our annotation is not to propose a novel benchmark for the evaluation community (hence only 100 dialogues), but as an artefact to highlight the limitations of current datasets being used to benchmark automatic dialogue evaluation. Nevertheless, our annotations are based on generations that better approximate current chatbot capabilities. Furthermore, our analysis show that these dialogues still contain language and culture-agnostic issues that evaluators ought to be able to detect. As such, our annotations may be used as a compliment to current benchmarks, and most importantly, as an example for future annotation efforts.

\section{Ethical Considerations}

\paragraph{Expert Annotations} All annotators are fluent in English and graduate level professionals in the field of Computation Linguistics (two of which authors of this work) and volunteered to conduct the annotation. Notwithstanding the diverse backgrounds, the annotation may still contain biases in evaluation process. For instance, given the expertise of these annotators in this field, their assessment of quality might differ from other groups. A larger, more diverse pool of annotators may reduce this bias, which was not considered in this work due to its small scale.

\paragraph{Monolingual} As identified in the Limitation section, our work, despite arguing for multilingual and multicultural benchmarks, conducts its experimentation in English. Additionally, all annotators share similar western cultural background. As such, it's conclusions are biased towards the evaluation of English dialogues, which may not extend to other cultures, specifically non-western ones. For instance, high context cultures \citep{Hall1959language} privilege non-verbal methods of communication, which is typically not transcribed into text \citep{commstyle}. 

\section*{Acknowledgments}

This research was supported by the Portuguese Recovery and Resilience Plan through project C645008882-00000055 (Responsible.AI) and by national funds through \textit{Fundação para a Ciência e a Tecnologia} (FCT) with references PRT/BD/152198/2021 and DOI: 10.54499/UIDB/50021/2020.

\bibliography{anthology,custom}

\end{document}